# Discovering Operational Patterns Using Image-Based Convolutional Clustering and Composite Evaluation: A Case Study in Foundry Melting Processes

Zhipeng Ma *, Bo Nørregaard Jørgensen * and Zheng Grace Ma *

SDU Center for Energy Informatics, The Maersk Mc-Kinney Moller Institute, Faculty of Engineering, University of Southern Denmark, DK-5230 Odense, Denmark
* Correspondence: zhma@mmmi.sdu.dk (Z.M.); bnj@mmmi.sdu.dk (B.N.J.); zma@mmmi.sdu.dk (Z.G.M.)

**Abstract**

Industrial process monitoring increasingly relies on sensor-generated time-series data, yet the lack of labels, high variability, and operational noise make it difficult to extract meaningful patterns using conventional methods. Existing clustering techniques either rely on fixed distance metrics or deep models designed for static data, limiting their ability to handle dynamic, unstructured industrial sequences. Addressing this gap, this paper proposes a novel framework for unsupervised discovery of operational modes in univariate time-series data using image-based convolutional clustering with composite internal evaluation. The proposed framework improves upon existing approaches in three ways: (1) raw time-series sequences are transformed into grayscale matrix representations via overlapping sliding windows, allowing effective feature extraction using a deep convolutional autoencoder; (2) the framework integrates both soft and hard clustering outputs and refines the selection through a two-stage strategy; and (3) clustering performance is objectively evaluated by a newly developed composite score, $S_{eva}$, which combines normalized Silhouette, Calinski–Harabasz, and Davies–Bouldin indices. Applied to over 3900 furnace melting operations from a Nordic foundry, the method identifies seven explainable operational patterns, revealing significant differences in energy consumption, thermal dynamics, and production duration. Compared to classical and deep clustering baselines, the proposed approach achieves superior overall performance, greater robustness, and domain-aligned explainability. The framework addresses key challenges in unsupervised time-series analysis, such as sequence irregularity, overlapping modes, and metric inconsistency, and provides a generalizable solution for data-driven diagnostics and energy optimization in industrial systems.

## 1. Introduction

The iron and steel industry is one of the most energy- and emission-intensive sectors, responsible for approximately 28.7% of global industrial $CO_2$ emissions in 2023 [1,2]. Foundries, in particular, face increasing pressure to improve energy efficiency while maintaining production reliability. Within these operations, furnace melting alone accounts for over half of total electricity consumption [3], making it a key target for data-driven optimization.

Modern foundries generate large volumes of time-series data from sensors monitoring temperature, power, and cycle durations. However, these data are typically unlabeled, nonstationary, and noisy, limiting the effectiveness of rule-based or supervised learning approaches. Unsupervised clustering of time-series data has therefore emerged as a promising strategy for uncovering latent operational modes, diagnosing inefficiencies, and enabling smart manufacturing [4,5]. Nonetheless, industrial time-series clustering remains challenging due to variable sequence lengths, overlapping regimes, and the absence of ground truth for model evaluation.

Classical time-series clustering methods such as k-means-DTW [6], k-shape [7], and hierarchical clustering [8] rely on handcrafted features or fixed distance metrics, restricting their applicability in complex industrial settings. More recent deep clustering approaches, such as DEC [9], IDEC [10], DTC [11], and EDESC [12], integrate representation learning with clustering. While effective on static or image data, these models struggle to capture localized temporal structure in noisy time series and typically rely on a single clustering paradigm, leading to unstable performance in the presence of ambiguous operational patterns.

To address these challenges, this paper proposes a novel deep clustering framework called TS-IDEC (Time-Series Image-based Deep Embedded Clustering) tailored to industrial time-series data. TS-IDEC introduces a hybrid pipeline that first transforms univariate sequences into grayscale images using an overlapping sliding window, enabling the use of convolutional feature extraction. This transformation bridges the gap between temporal and spatial representations, allowing a deep convolutional autoencoder (DCAE) to learn both localized and global structural features. Building on the IDEC model [10], TS-IDEC integrates both soft probabilistic clustering and hard centroid-based clustering, which are evaluated jointly to improve robustness and reliability.

To support unsupervised model selection and evaluation, this paper proposes a composite internal evaluation metric, $S_{eva}$, which combines normalized and ranked forms of three widely used clustering indices: the Silhouette Score [13], Calinski–Harabasz Index [14], and Davies–Bouldin Index [15]. This score mitigates inconsistencies among individual metrics. A two-stage evaluation process uses this score to rank dual-mode clustering outputs and select the best solution.

This paper validates TS-IDEC on a real-world case study involving 3900 univariate temperature profiles from an industrial induction furnace in a Nordic foundry. The framework successfully discovers meaningful operational modes in energy-intensive melting processes, demonstrating both scientific and practical relevance. More broadly, TS-IDEC addresses challenges common to many industrial and cyber-physical systems, including smart grids, chemical plants, and manufacturing lines, where data are noisy, variable, and unlabeled.

The scientific contributions of this work are as follows:

- A novel unsupervised clustering framework combining image-based sequence transformation with convolutional representation learning;
- A dual-mode clustering strategy that integrates soft and hard assignments for improved robustness;
- A composite evaluation metric for reliable internal validation without ground truth;
- Application to real industrial data, demonstrating discovery of operational modes in energy-intensive furnace operations.

The remainder of the paper is structured as follows. Section 2 reviews background and related work; Section 3 introduces the TS-IDEC framework; Section 4 presents the foundry case study; Section 5 reports experimental results, including benchmarks, robustness tests, and ablations; Section 6 discusses implications and situates findings within the literature; and Section 7 concludes with a summary, limitations, and directions for future research.

## 2. Background and Related Work

This section outlines the technical and domain context relevant to our proposed clustering framework. It begins by positioning the industrial motivation through the lens of energy-intensive foundry operations. It then surveys the methodological landscape of time-series clustering, focusing on both traditional and deep learning-based approaches. Finally, it discusses internal clustering evaluation metrics, highlighting their limitations in unsupervised settings. These foundations help contextualize the problem space and clarify the contributions of this work.

### 2.1. Industrial Context: Foundry Operations and Energy Challenges

The iron and steel industry remains one of the most energy- and emission-intensive sectors globally, accounting for 28.71% of industrial greenhouse gas emissions in 2023 [1,2]. Despite advances in electrification and furnace technologies, the sector remains behind the International Energy Agency's Net Zero by 2050 trajectory [16].

Foundries are critical to supply chains in automotive, infrastructure, and machinery, yet their operations are highly energy-intensive. Furnace melting alone accounts for about 55% of total foundry energy use [3]. With rising environmental pressures, particularly in Europe [17], improving the efficiency and flexibility of melting processes has become a strategic imperative [18].

A persistent challenge is that process optimization in foundries still depends largely on operator expertise and tacit knowledge. Melting patterns are diverse and dynamic, and standardized methods for assessing furnace performance are lacking. Traditional heuristic approaches fail to generalize across time, materials, or operating conditions. This gap underscores the need for machine learning approaches capable of modeling nonlinear temporal dynamics and uncovering latent operational modes. Unsupervised clustering, in particular, offers a promising direction by enabling the discovery of hidden process patterns directly from high-frequency industrial data [19,20].

### 2.2. Time-Series Clustering for Industrial Applications

### 2.2.1. Traditional Time-Series Data Clustering Approaches

Time-series clustering groups sequences based on temporal similarity in shape, amplitude, or dynamics. In industrial analytics, it is widely applied to identify operational modes, detect faults, and reduce data dimensionality [21]. Traditional approaches combine hand-engineered features with classical clustering algorithms such as k-means, agglomerative clustering, and Gaussian Mixture Models [4,5,22].

A central challenge is defining similarity measures for time-series data. Dynamic Time Warping (DTW) [23] remains foundational for aligning sequences of varying length and speed, with extensions such as soft-DTW [24] and weighted DTW [6] improving smoothness and noise robustness. DTW has been applied in process control and machine diagnostics, but its quadratic time complexity limits scalability without approximations.

To improve efficiency, k-shape [7] adapts k-means with a shape-based distance metric derived from normalized cross-correlation, enabling clustering by global trends while remaining invariant to amplitude shifts. However, k-shape and similar methods require length normalization and often fail to capture nonlinear temporal dynamics or local distortions.

More broadly, traditional clustering assumes static, linearly separable clusters and relies heavily on manual feature engineering (e.g., peak detection, signal derivatives, domain rules). These constraints limit generalizability and reduce effectiveness in industrial environments, where data are noisy, nonstationary, and unlabeled [25,26].

These limitations have driven a shift toward deep learning–based representation learning approaches, which jointly learn feature hierarchies and clustering structures directly from raw data. The next section reviews this emerging direction.

### 2.2.2. Deep Learning-Based Clustering

Deep clustering algorithms have gained prominence for their ability to jointly learn latent representations and cluster assignments in high-dimensional, noisy datasets. Among the most widely adopted are autoencoder-based models such as Deep Embedded Clustering (DEC) [9] and Improved DEC (IDEC) [10], which unify reconstruction and clustering losses within a single optimization loop. These frameworks enable efficient unsupervised learning pipelines and have been applied across diverse static data domains.

To capture local patterns in data with spatial or sequential structure, convolutional autoencoder (CAE) architectures have been introduced, particularly in image clustering [27]. In this study, we adapt a convolutional architecture to time-series data by transforming sequences into two-dimensional images, thereby enabling localized temporal feature extraction through convolutional filters.

Beyond autoencoders, other paradigms have emerged. GAN-based approaches, such as ClusterGAN and InfoGAN, exploit generative learning but often suffer from training instability. Subspace-based methods, including EDESC [12], provide scalability by clustering in learned latent spaces, but lack mechanisms for sequential modeling. More recently, topology-aware frameworks such as DEETO [28] have sought to preserve geometric structures in embedding spaces, though their added complexity limits practical deployment in industrial settings.

A summary of these categories, along with their advantages, limitations, and relevance to industrial time-series data, is provided in Table 1.

**Table 1.** Comparison of Deep Clustering Approaches for Industrial Time-Series Data.

| Clustering Approach | Representative Methods | Strengths | Limitations | Suitability for Industrial Time-Series |
|---|---|---|---|---|
| Autoencoder-based | DEC [9]; IDEC [10] | Joint feature learning and clustering; low-dimensional embeddings | Often assumes independently and identically distributed data; ignores temporal dependencies | Moderate: needs adaptation for sequence learning |
| Convolutional Autoencoder | DCAE [27]; TS-IDEC (this study) | Learns localized spatial/temporal patterns; scalable to image-structured input | Performance depends on image transformation quality | High: effective after converting the time-series to image representations |
| GAN-based | ClusterGAN [29]; InfoGAN [30] | Captures complex distributions; good for generative synthesis | Training instability | Low: rarely used for time-series due to mode collapse and training cost |
| Subspace clustering | EDESC [12]; Deep Subspace Clustering [31] | Models latent subspaces; scalable; suitable for high-dimensional data | Less effective when a strong temporal order exists | Moderate: best for multivariate sensor data, not univariate signals |
| Topology-aware | DEETO [28]; DA-Net [32] | Preserves geometric and topological structure in the latent space | High complexity; difficult to scale; sensitive to hyperparameters | Low to Moderate: promising for structured data but immature for industry |

These categories highlight the trade-offs between expressiveness and scalability in deep clustering for industrial analytics. Autoencoder-based methods provide strong baselines but largely neglect temporal dependencies. GAN- and topology-aware models increase modeling capacity yet remain difficult to tune and deploy. Subspace methods are efficient but disregard the sequential structure. Convolutional autoencoders, as adopted in the proposed TS-IDEC framework, offer a balanced solution: they enable robust feature extraction from transformed time-series inputs while remaining computationally practical. When combined with domain-aware clustering and evaluation strategies, this architecture forms the methodological foundation of our approach.

### 2.2.3. Clustering in Industrial Time-Series Contexts

Industrial time-series data are central to monitoring and optimizing performance in manufacturing, energy, transportation, and process industries. Generated by sensors and control systems, these sequences are typically nonstationary, multiscale, and noisy, often exhibiting abrupt regime shifts or operator interventions [8]. Clustering such data can uncover latent operational modes, support anomaly detection, and enable predictive maintenance and energy efficiency strategies.

Several studies have demonstrated the utility of clustering in industrial contexts. For example, hierarchical clustering has been used to group production runs in chemical vapor deposition processes, improving model calibration [33]. Gaussian mixture models have segmented turbine operational stages, aligning clusters with mechanical transitions [22]. Structure entropy–based clustering has enhanced stability in online monitoring of nonlinear processes under changing conditions [8].

Despite these advances, most prior work relies on traditional algorithms or hand-engineered features. These methods assume linear separability or static data structures and often fail under the complexity of modern industrial environments, where sequences vary in length, exhibit overlapping regimes, and reflect operator-induced variability [25,26].

Moreover, deep clustering methods have rarely been adapted to the unique characteristics of industrial time series. While such methods have shown promise in healthcare [34], energy systems [35], and power plant performance [36], few studies address heavy manufacturing domains such as metal foundries, where data are unlabeled, noisy, and characterized by complex thermal and energy dynamics. General-purpose models, such as DEC or DTC, often fail to capture temporal variability, contextual dependencies, and the level of explainability required by domain experts.

This gap motivates the need for purpose-built clustering frameworks that:

- Extract temporal features relevant to industrial signals;
- Remain robust to variable-length and noisy sequences;
- Integrate soft and hard clustering for stability and reliability;
- Incorporate internal evaluation tools suitable for unsupervised, label-free domains.

### 2.3. Internal Clustering Evaluation

In unsupervised learning, especially in industrial contexts where labels are unavailable, internal evaluation metrics are critical for assessing cluster quality. These measures evaluate cohesion (intra-cluster similarity) and separation (inter-cluster dissimilarity) based solely on data structure.

Three commonly used internal metrics are:

- Silhouette Score (SIL) [13]: Measures the mean intra-cluster distance vs. the nearest-cluster distance, normalized to [−1, 1]. A higher score indicates better-defined clusters.

- Calinski–Harabasz Index (CH) [14]: Computes the ratio of between-cluster dispersion to within-cluster dispersion. A higher CH score suggests clearer separation.
- Davies–Bouldin Index (DB) [15]: Evaluates cluster similarity by comparing each cluster with its most similar neighbor. Lower values indicate more distinct clusters.

Although widely adopted, these metrics often yield conflicting results. For example, CH tends to favor larger numbers of clusters in high-dimensional spaces, while SIL penalizes over-segmentation by rewarding cohesion. DB is sensitive to outliers and non-uniform cluster spreads. Such inconsistencies are particularly pronounced in time-series clustering, where temporal misalignment and data sparsity distort distance-based assumptions. Divergent outcomes are common: CH may report high values when sequences differ in amplitude or duration, even if temporal separation is weak, whereas SIL penalizes overlap caused by time warping. Similarly, CH rewards global separation, while DB emphasizes local overlap, leading to inconsistent optima. SIL further degrades in high-dimensional feature spaces generated by windowed or encoded time-series data.

Several additional metrics, such as the Dunn index and entropy-based criteria, have also been explored, but they tend to suffer from instability in noisy environments or are difficult to interpret in industrial domains.

Table 2 summarizes and contrasts these internal metrics with respect to their assumptions, strengths, and limitations in the context of time-series clustering.

**Table 2.** Comparison of Internal Clustering Evaluation Metrics.

| Metric | Core Idea | Output Range | Strengths | Limitations | Suitability for Time-Series Clustering |
|---|---|---|---|---|---|
| Silhouette Score (SIL) | Balance between cohesion and separation | [−1, 1] | Intuitive; interpretable; shape-agnostic | Suffers in high dimensions; may penalize non-convex clusters | Moderate: interpretable but sensitive to time warping |
| Calinski–Harabasz (CH) | Ratio of between-to within-cluster variance | [0, ∞) | Computationally efficient; works well for spherical clusters | Biased toward higher cluster counts; assumes isotropic structure | Moderate: fast but sensitive to density imbalance |
| Davies–Bouldin (DB) | Average similarity between each cluster and its nearest | [0, ∞) | Low values favor compact, well-separated clusters | Sensitive to outliers and centroid instability | Moderate: useful but unreliable with noise |
| Dunn Index | Ratio of minimum inter-cluster to maximum intra-cluster | [0, ∞) | Encourages compactness and separation | Highly sensitive to noise and outliers; rarely stable | Low: rarely used in noisy industrial settings |
| Entropy-based indices | Based on cluster label uncertainty | N/A | Captures uncertainty and distributional overlap | Requires discretization; less interpretable | Low: useful in hybrid methods or fuzzy clustering |
| Composite Scores | Weighted/normalized combination of metrics | [0, 1] (typical) | Mitigates individual biases; balances trade-offs | Requires normalization and weighting schemes; no standard method | High: best suited for unsupervised industrial analytics |

To address these issues, several studies have explored composite evaluation frameworks. A comparative study of over 30 indices concluded that no single metric consistently performs well across data types and recommended combining multiple measures [37]. More recent work [38] has adopted ensemble strategies, including rank aggregation, majority voting, and normalized weighted averages. While these approaches improve robustness, challenges remain: scale mismatches between metrics, sensitivity to outliers, and lack of adaptation to noisy, high-dimensional time-series data. These limitations underscore the need for more reliable, explainable, and domain-specific evaluation methods, motivating the composite metric proposed in this study.

# 3. Proposed Framework: Deep Clustering with Composite Evaluation for Industrial Time-Series

This section presents the proposed unsupervised learning framework for clustering industrial time-series data, with a focus on modeling operational patterns in foundry processes. The framework, termed Time-Series Image-based Deep Embedded Clustering (TS-IDEC), combines temporal transformation, deep feature extraction, and a dual-mode clustering mechanism, enabling robust and reliable clustering of univariate process signals without the need for labels. To address the challenges of noisy, variable-length industrial time series, the framework begins by transforming raw sequences into two-dimensional representations using an overlapping sliding window technique. These transformed inputs are then processed by a deep convolutional autoencoder, which learns compact latent representations optimized jointly for reconstruction and clustering objectives. Both soft and hard clustering results are derived from the latent space and evaluated using a composite internal metric, designed to resolve conflicting signals from standard clustering quality indices.

## 3.1. TS-IDEC Architecture and Design Principles

The proposed Time-Series Image-based Deep Embedded Clustering (TS-IDEC) framework is designed to address three core challenges in clustering industrial time-series data: (i) capturing localized temporal patterns within noisy sequences, (ii) improving clustering robustness in the absence of ground truth, and (iii) integrating explainable model evaluation and selection into the unsupervised learning pipeline. TS-IDEC builds upon the IDEC algorithm developed by [10], but extends it in multiple critical ways to support its application to industrial contexts.

First, to better preserve temporal structure and local pattern continuity, each univariate time series is transformed into a two-dimensional grayscale image using an overlapping sliding window technique, as described in Section 3.2. This transformation enables the use of convolutional operations that are sensitive to spatial and temporal locality, thus enhancing the network's ability to extract meaningful sequence features.

Second, the fully connected layers in the original IDEC architecture are replaced with a deep convolutional autoencoder (DCAE). The encoder component is composed of convolutional and pooling layers, while the decoder consists of deconvolutional and upsampling layers. This architectural shift enables the network to leverage local translation-invariant features and scale more effectively with larger datasets, both of which are essential for analyzing long-duration operational sequences from industrial processes.

Third, TS-IDEC introduces a dual-mode clustering strategy. A soft clustering layer, based on Student's t-distribution kernel, is used during training to refine latent representations while preserving cluster structure [9]. In parallel, a hard clustering approach, based on the k-means algorithm, is applied to the same latent space after training to improve cluster separation. These two outputs, denoted as $C_1$ (soft) and $C_2$ (hard), are subsequently evaluated using a robust internal metric, and the better-performing result is selected as the final clustering solution (see Section 3.5).

After defining the architectural components, it is important to clarify how the training and clustering output selection are performed. TS-IDEC is optimized by minimizing a joint loss function composed of a reconstruction loss, denoted $L_{rec}$, and a clustering loss, denoted $L_{cl}$. The reconstruction loss ensures that the autoencoder learns structure-preserving latent representations from the grayscale image input, while the clustering loss refines those representations to improve cluster compactness and separation. The combined loss function is optimized using stochastic gradient descent. Upon convergence, the model produces two candidate clustering results: a soft assignment output $C_1$ derived from the probabilistic clustering layer, and a hard clustering output $C_2$ generated by applying the k-means algorithm to the latent space. A two-stage evaluation procedure, based on qualitative consistency and a composite internal metric (see Section 3.5), is used to select the better-performing result. The final output is denoted as $C_{best}$, representing the clustering result that balances accuracy and reliability.

The complete architecture of TS-IDEC is illustrated in Figure 1. The input time series $x_0$ is first transformed into a grayscale matrix $\widehat{x_1}$, which serves as the input to the convolutional encoder. The encoder maps this input to a latent representation $Z \in R^{128}$, which is simultaneously used for reconstruction through the decoder and for clustering through both soft and hard assignment pathways. The training process jointly minimizes a reconstruction loss and a clustering loss, ensuring that latent representations are both compact and cluster-discriminative.

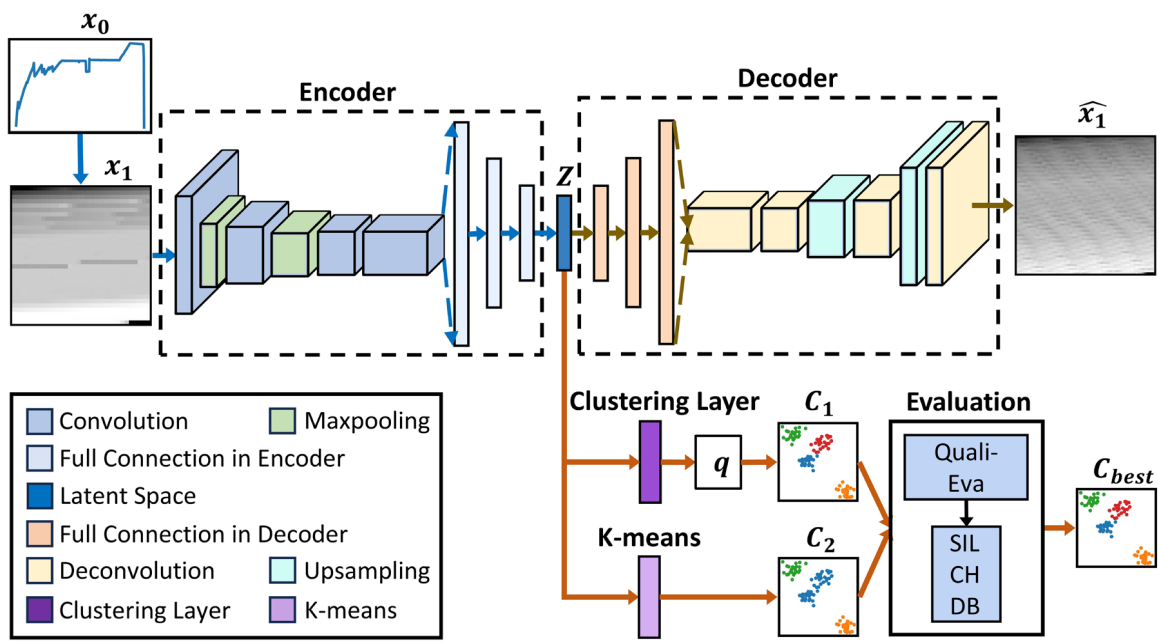

**Figure 1.** Architecture of the TS-IDEC network. The pipeline includes overlapping sliding window transformation, convolutional encoder, latent space extraction, decoder, and dual clustering modules. Color-coded blocks denote functional components.

This architectural design makes TS-IDEC particularly suitable for industrial time-series clustering, where data are often unlabeled, irregular, and sensitive to small operational variations. The integration of time-series-aware input transformation, deep spatial-temporal feature learning, and dual clustering evaluation ensures the framework provides a robust and scalable solution for process pattern discovery.

### 3.2. Temporal Transformation Via Overlapping Sliding Windows

Time-series data from industrial operations often exhibit variable sequence lengths, local fluctuations, and nonstationary behavior, making direct clustering difficult. To enhance pattern recognition and preserve temporal dependencies, TS-IDEC first transforms each univariate time series into a structured, image-like representation using an overlapping sliding window technique. This preprocessing step enhances spatial coherence, facilitates localized feature extraction through convolution, and improves robustness to noise.

Let the original univariate time series be denoted as $x_0 \in R^N$, where $N$ is the number of time points. The overlapping sliding window segments $x_0$ into $n_s$-length windows with an overlap of $n_o$, where ($n_o < n_s$). These windows are stacked as rows to form a two-dimensional matrix $x_1 \in R^{n_s \times n_s}$, where both the number of rows and columns are equal to the window size. This results in a grayscale image-like structure suitable for convolutional processing.

The total number of values $N_1$ required to construct this matrix is given by:

$$N_1 = n_s + (n_s - 1)(n_s - n_o) \tag{1}$$

To ensure that all $N$ values in time series are utilized, appropriate values for $n_s$ and $n_o$ should be selected to satisfy the condition $N_1 \geq N$. If $N_1 > N$, padding of $N_1 - N$ values can be applied as a preprocessing step for the time series.

An illustration of this transformation process is shown in Figure 2. The original time series $x_0$ is segmented into overlapping chunks, each contributing a row to the matrix $x_1$. Colored boxes in the left-hand plot represent individual sliding windows, while their corresponding rows are mapped to the grayscale image on the right. The transformation preserves both short-term temporal patterns within windows (horizontal axis) and transitions across windows (vertical axis), thus enabling dual-perspective feature extraction during learning.

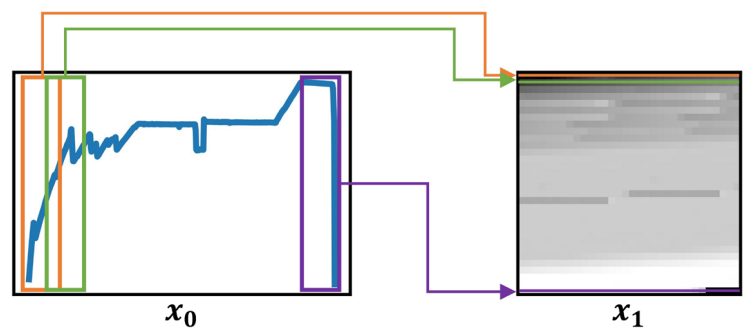

**Figure 2.** Example of overlapping sliding window transformation. The original univariate time series $x_0$ is segmented into overlapping windows (**left**), which are stacked into a grayscale matrix $x_1$ (**right**) suitable for convolutional feature extraction.

This encoding of temporal structure helps capture fine-grained operational transitions and long-range dependencies—two common challenges in real-world process monitoring [21,26]. Moreover, overlapping improves generalization by generating multiple slightly varied training instances from the same base sequence, while minimizing the risk of omitting important transitions between operational states.

By converting time-domain signals into structured spatial representations, this transformation bridges the gap between time-series modeling and image-based convolutional learning. The resulting matrices serve as input to the encoder in TS-IDEC, enabling the downstream architecture to exploit spatial locality and structural regularities that would be difficult to learn from raw sequences directly.

### 3.3. Deep Convolutional Autoencoder and Clustering Mechanism

The core of the TS-IDEC framework is a deep convolutional autoencoder (DCAE), which is responsible for learning compact and informative latent representations of time-series input transformed into grayscale matrices. As shown in Figure 1, the DCAE comprises three main components: an encoder, a latent space, and a decoder. This structure allows the model to learn spatial-temporal features that reflect both local and global process dynamics, which are essential for effective clustering.

Let the transformed 2D input be denoted $x_1 \in R^{n_s \times n_s}$. The encoder maps each input sample $x_i$ to a latent representation $Z \in \mathrm{R}^{128}$ through a series of convolutional and pooling layers, defining a nonlinear transformation $f_W: x_i \longrightarrow z_i$. This latent space $Z$ preserves the most salient features of the time-series input while reducing its dimensionality. The decoder applies the inverse transformation $g_{W'}: z_i \longrightarrow \hat{x}_i$ using deconvolution and up-sampling layers to reconstruct the original input. The reconstruction ensures that the latent features retain meaningful temporal structure and are not arbitrarily compressed.

The architectural parameters of the DCAE were selected through iterative testing to balance reconstruction fidelity with clustering effectiveness. The final configuration is summarized in Table A1, which lists each layer's type, kernel size, stride, padding, and activation function. The latent dimensionality was fixed at 128 based on experimental evaluation of compression quality and clustering performance.

Once the latent space $Z$ is learned, clustering is performed using a dual-mode strategy. The first mode is a soft clustering mechanism, implemented through a Student's $t$-distribution kernel as proposed by [9]. Given the initial $k$ cluster centroids $\{\mu_j\}(j = 1, \dots k)$ obtained through a simple clustering algorithm such $k$-means clustering and the data points $z_i$ from the latent space $Z$, the probability of assigning a sample $z_i$ to cluster $j$, with centroid $\mu_j$, is computed as:

$$q_{i,j} = \frac{(1 + \left\| z_i - \mu_j \right\|^2 / \alpha)^{-\frac{\alpha+1}{2}}}{\sum_{j'} (1 + \left\| z_i - \mu_{j'} \right\|^2 / \alpha)^{-\frac{\alpha+1}{2}}} \tag{2}$$

where $\alpha$ refers to the degree-of-freedom parameter of the t-distribution, $q_{i,j}$ denotes the probability of assigning sample $i$ to cluster $j$, and $\|\cdot\|$ denotes the Euclidean norm. This soft assignment matrix $Q$ is used to refine cluster assignments during training and contributes directly to the clustering loss (see Section 3.4). Each sample is assigned to its most probable cluster according to $\arg\max_j q_{i,j}$, forming the soft clustering output $C_1$.

The second mode is hard clustering, performed by applying the standard k-means algorithm directly to the latent space $Z$ after training. This produces a second candidate output $C_2$, which is typically more stable for deployment in industrial decision systems.

Next, to evaluate the clustering quality of $C_1$ and $C_2$, a two-step evaluation approach is employed. The first stage involves qualitative assessment. In some experiments with complex datasets, the data points may not be well separated. For example, the soft clustering result $C_1$ may consist of only one cluster, or a hard clustering result $C_2$ may contain certain clusters with an extremely small number of data points. If one of these results is suboptimal based on the predefined clustering settings and requirements, the other is directly selected as the superior clustering result, denoted as $C_{best}$. If both exhibit similar performance in the qualitative assessment, a subsequent quantitative evaluation is performed. In this step, to determine the most appropriate final clustering output, both $C_1$ and $C_2$ are evaluated using a composite evaluation score integrating the Silhouette Score (SIL), Calinski–Harabasz Index (CH), and Davies–Bouldin Index (DB), as discussed in Section 3.5. The better-performing output is selected as $C_{best}$. This two-stage evaluation ensures that the final clustering solution is both quantitatively sound and domain-explainable, which is essential for decision support in industrial applications.

This architecture can improve robustness to noise, enhance pattern separation, and explain operational modes for time-series data in complex industrial environments due to the joint leverage of the strength of convolutional representation learning and dual clustering comparison.

### 3.4. Joint Optimization of Reconstruction and Clustering Objectives

The TS-IDEC framework is trained through a joint optimization of two objective functions: the reconstruction loss $L_{rec}$ and the clustering loss $L_{cl}$. This approach ensures that the learned latent representations are not only able to reconstruct the input data accurately but are also discriminative for clustering. This dual-objective formulation enables the latent space to reflect both the structure of the time-series inputs and the underlying grouping patterns.

The total loss function $L$ is defined as a weighted sum of the reconstruction loss $L_{rec}$ and the clustering loss $L_{cl}$, following the formulation in [10]:

$$L = L_{rec} + \gamma L_{cl}, \tag{3}$$

where $\gamma > 0$ is a hyperparameter that balances the relative importance of clustering against reconstruction during training.

The reconstruction loss $L_{rec}$ is measured using the mean squared error (MSE) between the input matrix $x_i$ and its reconstructed counterpart $\hat{x}_i$:

$$L_{rec} = \sum_{i=1}^{n} \|x_i - \hat{x}_i\|^2 \tag{4}$$

This loss ensures that the latent representation retains enough temporal and structural information to reconstruct the original sequence with minimal distortion.

The clustering loss, first introduced in [9], is defined as the Kullback–Leibler (KL) divergence between the soft assignment distribution $Q = [q_{i,j}]$ and the target distribution $P = [p_{i,j}]$, which is constructed to sharpen the cluster assignments and emphasize points with high-confidence predictions:

$$L_{cl} = \mathrm{KL}(P\|Q) = \sum_i \sum_j p_{i,j} \log \frac{p_{i,j}}{q_{i,j}}, \tag{5}$$

where $q_{i,j}$ is defined in Equation (2) as the probability of assigning a sample $i$ to cluster $j$, and $p_{i,j}$ is the auxiliary distribution presented in Equation (6):

$$p_{i,j} = \frac{q_{i,j}^2/f_j}{\sum_{j'} q_{i,j'}^2/f_{j'}}, \tag{6}$$

where $f_j = \sum_i q_{i,j}$ denotes the frequency of samples assigned to the cluster $j$. This target distribution serves three purposes: it enhances cluster purity by emphasizing confident assignments, balances cluster influence by normalizing frequencies, and helps refine cluster boundaries during optimization [9].

The loss is jointly optimized using the Adam optimizer [39], which updates both the model weights and the cluster centroids. The gradients of the clustering loss with respect to the latent vectors $z_i$ and the centroids $\mu_j$ are computed as follows:

$$\frac{\partial L_{cl}}{\partial z_i} = \frac{\alpha+1}{\alpha} \sum_j (1 + \frac{\|z_i - \mu_j\|^2}{\alpha})^{-1} \times (p_{i,j} - q_{i,j})(z_i - \mu_j) \tag{7}$$

$$\frac{\partial L_{cl}}{\partial \mu_j} = -\frac{\alpha+1}{\alpha} \sum_j (1 + \frac{\|z_i - \mu_j\|^2}{\alpha})^{-1} \times (p_{i,j} - q_{i,j})(z_i - \mu_j) \tag{8}$$

Given a mini batch with $m$ samples and the learning rate $\lambda$, each $\mu_j$ is updated as follows:

$$\mu_j = \mu_j - \frac{\lambda}{m} \sum_{i=1}^{m} \frac{\partial L_{cl}}{\partial \mu_j} \tag{9}$$

The decoder weights $W'$ are updated as follows:

$$W' = W' - \frac{\lambda}{m} \sum_{i=1}^{m} \frac{\partial L_{rec}}{\partial W'} \qquad (10)$$

The encoder weights $W$ are updated as follows:

$$W = W - \frac{\lambda}{m} \sum_{i=1}^{m} \left( \frac{\partial L_{rec}}{\partial W} + \gamma \frac{\partial L_{cl}}{\partial W} \right) \qquad (11)$$

This optimization strategy ensures that the latent representations learned by the encoder are simultaneously reconstructive and conductive to effective clustering. The balance coefficient $\gamma$ is chosen empirically to avoid overfitting to either objective.

### 3.5. Composite Internal Evaluation Without Ground Truth

In the absence of labeled data, the evaluation of clustering performance relies on internal validation metrics that assess the compactness and separation of clusters based solely on the input data structure. While widely used, individual internal metrics often yield conflicting or unstable results, particularly when applied to high-dimensional, noisy, and irregular time-series data, which is typical in industrial contexts. To overcome this limitation, this paper proposes a composite internal evaluation strategy that integrates multiple standard metrics into a unified, normalized score, enabling robust and interpretable model selection.

This paper focuses on three widely used internal clustering indices: the Silhouette Score (SIL) [13], the Calinski–Harabasz Index (CH) [14], and the Davies–Bouldin Index (DB) [15]. These metrics capture complementary aspects of clustering quality—cohesion, separation, and compactness—but differ in scale, sensitivity to outliers, and response to cluster shape and density. As shown in earlier studies [37], no single metric performs consistently across all datasets, and metric-specific biases can distort evaluation outcomes in unsupervised settings.

To construct a robust evaluation score, this paper applies a two-stage normalization and aggregation procedure. First, each metric is rescaled using min–max normalization, with outlier effects mitigated through interquartile range (IQR) filtering. Then, each score is converted into a rank-based value, which reduces the influence of scale and irregular distributions. The final normalized score for each metric $i \in \{SIL, CH, DB\}$ is computed as the average of its min–max scaled and rank-normalized forms:

$$S_{norm}^{i} = (S_{minmax}^{i} + S_{rank}^{i})/2, \qquad (12)$$

where $S_{minmax}^{i}$ denotes the scores obtained after applying min-max scaling to the values with outliers removed, and $S_{rank}^{i}$ refers to the scores after applying rank normalization to the original evaluation scores, where $i$ corresponding to the three evaluation metrics.

Let $X$ denote the original vector of scores and $X'$ the same vector with outliers removed using IQR filtering. The min–max normalized score $S_{minmax}^{i}$ is computed differently depending on whether higher or lower values indicate better clustering:

$$S_{minmax}^{i} = \begin{cases} \dfrac{X' - \min(X')}{\max(X') - \min(X')} & i = SIL \ or \ CH \\ 1 - \dfrac{X' - \min(X')}{\max(X') - \min(X')} & i = DB \end{cases} \qquad (13)$$

The rank-normalized score $S_{rank}^{i}$ is defined as:

$$S_{rank}^{i} = \frac{\text{rank}(X) - 1}{N - 1} \qquad (14)$$

where $\text{rank}(\cdot)$ is the rank of $X$ in a sorted order and $N$ refers to the number of values in the dataset. If $i = SIL$ or $CH$, $\text{rank}(\cdot)$ is computed in ascending order, whereas if $i = DB$, $\text{rank}(\cdot)$ is computed in descending order.

Finally, the composite evaluation score $S_{eva}$ is calculated by averaging the normalized scores across all three metrics:

$$S_{eva} = (\sum_i S_{norm}^i)/3 \tag{15}$$

Based on the calculation in Equations (12)–(15), $S_{eva}$ is in the range $[0,1]$, and a higher value indicates a better clustering performance.

This composite score is used to select the final clustering result $C_{best}$ from the two candidate outputs produced by TS-IDEC. Specifically, if the soft clustering result both $C_1$ and the hard clustering result $C_2$ yield distinct scores, the result with the higher $S_{eva}$ is selected. Additionally, this composite score is also employed to compare the performance of the TS-IDEC algorithm with the baseline methods and determine the optimal number of clusters in the absence of ground truth in Section 5.

# 4. Case Study: Clustering Operational Patterns in Foundry Furnace Melting Processes

## 4.1. Industrial Process Context and Energy Characteristics

The data are generated from a large Danish foundry, which is one of the largest foundries in the Nordic countries. Figure 3 demonstrates the production flow of the foundry processes [1,40]. The small light blue rectangles within the larger rectangle, labeled with the respective stages, represent the facilities corresponding to different production stages. For instance, six induction furnaces are depicted. The solid blue rectangles indicate the movement of materials, while a circle with a cross signifies the completion of processing. Solid arrows denote the mandatory production flow, whereas dashed arrows represent optional production paths, as ladle preheating is not required in all cases. The induction furnaces are enclosed within a light gray rectangle with a dashed-line border, as expert insights from the facility indicate that these furnaces contribute to over half of the total electricity consumption.

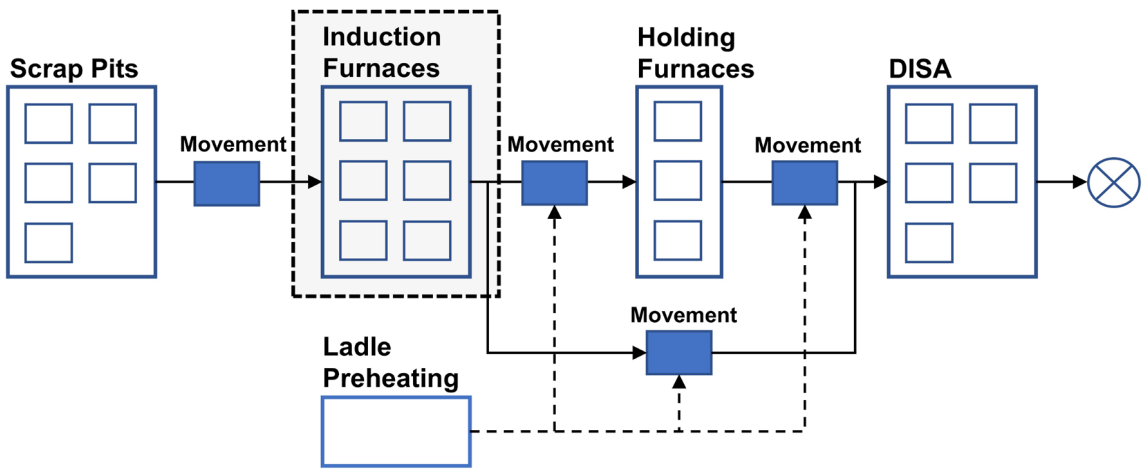

**Figure 3.** Industrial processing operations in the foundry.

The production process starts with the collection and categorization of raw materials from scrap metal pits. Following this, iron is fed into induction furnaces, where it is heated to a specified temperature while accounting for ferromagnetic losses. Once the molten metal reaches the required temperature, it is poured into a preheated transfer ladle. At this stage, alloying elements may be introduced to modify the composition, ensuring compatibility with the molten metal in the holding furnaces. The melt is subsequently transferred to a holding furnace, where it remains for a controlled duration. This marks the conclusion of the melting stage in the foundry process. The next phase, primary forming, then begins, shaping the material into various products.

For a comprehensive analysis of energy consumption patterns, it is advisable to focus on operational practices related to the induction furnaces. The electricity consumption in these furnaces is primarily driven by the need to increase the temperature to melt metal. Consequently, operational activities influencing electricity usage are reflected in the temperature dynamics within the furnace.

### 4.2. Data Acquisition, Segmentation, and Preprocessing

The dataset used in this study was collected from one of the six induction furnaces in the plant over a continuous monitoring period from 14 November 2022 to 30 April 2024, with temperature sampled at 10 s intervals. A visual overview of the selected furnace and sample temperature data is shown in Figure 4. Sub-Figure 4a displays a photo of the furnace, while sub-Figure 4b illustrates a typical melting operation profile, including heating, melting, and cooling stages.

In sub-Figure 4b, the blue curve represents the melting operation over time, while the red rectangular blocks highlight individual melting operations. Three operations are illustrated as examples. The figure demonstrates that different operations vary in their initial temperature, duration, and overall profile.

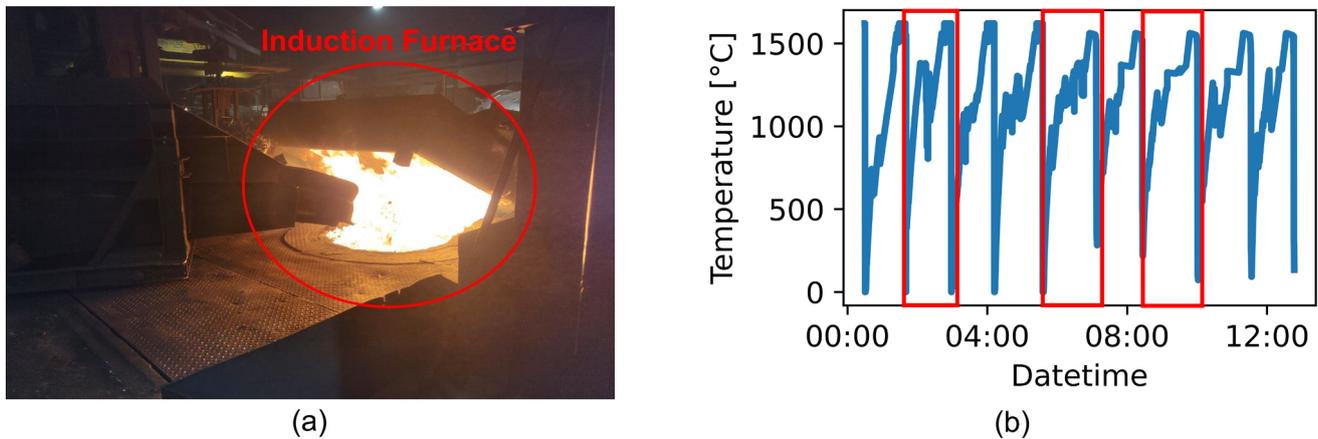

(a)                          (b)

**Figure 4.** (**a**) Selected induction furnace used in the case study. (**b**) Sample temperature profile of a complete melting operation, where the blue curve represents the melting operation over time, and the red boxes highlight individual melting cycles.

Each melting operation begins at ambient temperature and proceeds through a heating phase until the metal reaches the required melting point. Following this, the molten metal is poured, and the furnace is cooled in preparation for the next cycle. This pattern repeats continuously, with each operation cycle forming a distinct univariate time series. A total of 3927 melting operations are identified during the monitoring period. No missing values are found in the collected melting data.

To construct these individual sequences, this paper developed a segmentation pipeline using domain-specific rules based on temperature thresholds and time-based heuristics. Each sequence is treated as a self-contained melting cycle, starting from preheating and ending after post-melt cooling. These time series exhibit significant noise, and occasional interruptions due to operator interventions (e.g., mid-process alloying or furnace door opening), making the dataset highly representative of real-world industrial complexity.

To enable input into the TS-IDEC framework, each raw time series was resampled to a fixed length of 497 points, corresponding approximately to the median operation duration. This was achieved via linear interpolation and aggregation, ensuring that key temporal patterns were preserved while enabling batch processing and direct comparison across sequences. Consequently, each input sequence is one-dimensional, consisting solely of temperature values, with a standardized length of 497.

This preprocessed dataset thus serves as the input to the clustering algorithm, allowing TS-IDEC to discover latent operational modes based on the shape and dynamics of melting temperature curves. The diversity, noise, and size of the dataset make it well-suited for evaluating the performance and generalizability of deep unsupervised clustering methods.

## 5. Experimental Evaluation and Analysis

This section presents a comprehensive evaluation of the proposed TS-IDEC framework through benchmark comparisons, domain-specific clustering analysis, robustness testing, and ablation studies. The experiments aim to (i) assess TS-IDEC's performance relative to classical and deep clustering methods, (ii) examine the quality and explainability of furnace operational clusters, (iii) validate the framework's stability under parameter variation, and (iv) justify the architectural and evaluation design choices through controlled ablations. All experiments are conducted on real-world univariate temperature time-series data collected from an industrial foundry, as described in Section 4.

### 5.1. Experimental Setup and Baselines

To evaluate the performance of the proposed TS-IDEC framework, this paper conducted a series of comparative experiments against classical and deep learning-based time-series clustering methods. These experiments were designed to assess clustering performance on the furnace melting dataset described in Section 4. The dataset comprises 3927 univariate temperature time series, each representing a complete melting operation.

All time-series instances were initially of variable length due to differing melting durations. To ensure consistency and compatibility with all clustering models, each sequence was resampled to a fixed length of 497 time steps using linear interpolation and aggregation. This length corresponds approximately to the median duration across all melting operations, enabling batch processing without excessive truncation or padding. The preprocessed sequences serve as the input to both classical and deep clustering baselines.

The proposed TS-IDEC was compared against seven clustering methods, as shown in Table 3. The k-Shape, k-Means-DTW, and k-Means-SoftDTW methods were implemented using the tslearn library (version 0.6.3) [41], while all deep clustering models, including TS-IDEC, were implemented in PyTorch 2.3.1 [42].

**Table 3.** Overview of Time Series Clustering Algorithms Used for Benchmarking.

| Algorithm | Reference | Description |
|-----------|-----------|-------------|
| k-Shape | [7] | A shape-based clustering algorithm using normalized cross-correlation. |
| k-Means-DTW | [6] | Classical k-means applied with Dynamic Time Warping (DTW) distance. |
| k-Means-SoftDTW | [24] | A differentiable version of DTW that allows smooth optimization. |
| IDEC | [10] | A deep-embedded clustering method using fully connected autoencoders and soft assignment. |
| IDEC-conv1D | Adapted from [10] | A modified version of IDEC using a 1D convolutional encoder to better capture temporal patterns. |

| DTC | [11] | Deep Temporal Clustering that integrates sequential feature learning and clustering in an end-to-end framework. |
| EDESC | [12] | A deep subspace clustering model designed for high-dimensional data, suitable for industrial applications. |

For TS-IDEC, the input sequences were first transformed into 32 × 32 grayscale matrices using the overlapping sliding window method described in Section 3.2. The convolutional autoencoder architecture followed the configuration shown in Table A1. Key hyperparameters were tuned based on a series of preliminary experiments:

- Learning rate: Chosen from {0.05, 0.01, 0.001, 0.0005}, with 0.001 yielding stable convergence.
- Batch size: Tested from {8, 16, 32, 64}; 32 was selected for its balance of efficiency and performance.
- Weighting coefficient $\lambda$: Set to 0.5 to balance reconstruction and clustering losses.
- Epochs: Set to 1000 to ensure convergence without overfitting.

All experiments were conducted on an NVIDIA GeForce RTX 3060 GPU using Python 3.11.7. Each model was trained five times with different random seeds, and results are reported as mean values with standard error.

The IDEC model used a traditional fully connected autoencoder with an encoder layer configuration of 256–128–64 neurons. Training was performed with a batch size of 8. To enhance temporal feature extraction, IDEC-conv1D replaced the encoder with a 1D convolutional architecture, detailed in Table A2 below.

This setup ensures a fair comparison between TS-IDEC and relevant baselines under consistent experimental conditions, allowing for rigorous assessment of clustering effectiveness, model robustness, and architectural design choices.

Moreover, more advanced sequence models directly on raw 1D sequences, including Temporal Convolutional Networks (TCNs) and Transformers, are experimented. However, their performance in this industrial setting was significantly lower than the reported baselines. For this reason, they are not involved in the benchmark table.

### 5.2. Benchmark Comparison: TS-IDEC vs. Baselines

To assess the clustering performance of TS-IDEC, this paper conducted a comparative evaluation against seven baseline algorithms introduced in Section 5.1. All methods were tested on the standardized furnace melting dataset using the same experimental conditions. Performance was evaluated using the proposed composite internal metric $S_{eva}$, which integrates the Silhouette Score (SIL), Calinski–Harabasz Index (CH), and Davies–Bouldin Index (DB), as described in Section 3.5.

Each algorithm was allowed to search for the optimal number of clusters within the range representation $k \in [3, 16]$, and the best performing result based on $S_{eva}$ was reported. The evaluation scores are summarized in Table 4 and visualized in Figure 5.

**Table 4.** Normalized evaluation scores of different algorithms.

| Methods | n_Clusters | $S_{norm}^{SIL}$ | $S_{norm}^{CH}$ | $S_{norm}^{DB}$ | $S_{eva}$ |
|---|---|---|---|---|---|
| k-shape | 3 | **0.9883 ± 0.0** | 0.0820 ± 0.0 | **0.9714 ± 0.0** | 0.6850 ± 0.0 |
| k-means-dtw | 3 | 0.7085 ± 0.03708 | 0.5026 ± 0.03122 | 0.4586 ± 0.02966 | 0.5566 ± 0.03052 |
| k-means-softdtw | 4 | 0.6477 ± 0.03654 | 0.4418 ± 0.02555 | 0.4727 ± 0.01966 | 0.5207 ± 0.00938 |
| IDEC | 3 | 0.6209 ± 0.05454 | 0.9032 ± 0.05866 | 0.6882 ± 0.03852 | 0.7374 ± 0.04277 |
| IDEC-Conv1d | 3 | 0.6115 ± 0.01907 | **0.9166 ± 0.02443** | 0.5424 ± 0.05474 | 0.6902 ± 0.02199 |
| DTC | 3 | 0.5320 ± 0.03071 | 0.7496 ± 0.05288 | 0.5081 ± 0.03712 | 0.5966 ± 0.03233 |
| EDESC | 3 | 0.6922 ± 0.00395 | 0.8775 ± 0.00347 | 0.6815 ± 0.00397 | 0.7504 ± 0.00154 |
| TS-IDEC | 4 | 0.6494 ± 0.02740 | 0.8246 ± 0.05510 | 0.7830 ± 0.03607 | **0.7523 ± 0.00724** |

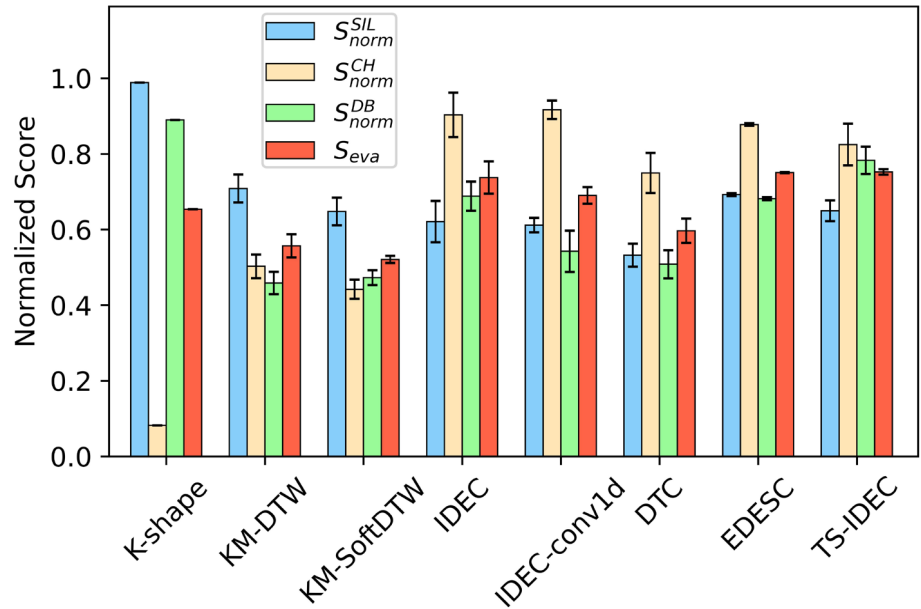

**Figure 5.** Normalized evaluation scores of different algorithms. The x-axis represents the names of algorithms, where "KM" denotes k-means. The y-axis shows the normalized scores ranging from 0 to 1. Each group of four bars corresponds to distinct evaluation metrics, as indicated in the legend, with error bars representing the standard error of repetitions.

The results in Table 4 and Figure 5 indicate that TS-IDEC achieves the highest overall performance based on the composite evaluation score $S_{eva}$ = 0.7523, outperforming all baselines, including state-of-the-art deep clustering methods such as IDEC and EDESC.

While TS-IDEC does not rank first in any of the individual normalized metrics, including $S_{norm}^{SIL}$, $S_{norm}^{CH}$, and $S_{norm}^{DB}$, it consistently ranks second or third across all three, highlighting its robustness and stability across diverse evaluation dimensions. In contrast, methods such as k-Shape and IDEC attain superior results on specific indices but show limitations in practical interpretability. For instance, k-Shape frequently merges sequences with distinct melting modes into a single cluster, yielding higher $S_{norm}^{SIL}$, but obscuring meaningful process differences. Likewise, IDEC often produces compact clusters that score well on $S_{norm}^{CH}$ yet collapse the majority of sequences into a single group. In contrast, TS-IDEC achieves a balanced performance across all indices while preserving operational distinctions, making it both robust and practically interpretable.

Among deep baselines, IDEC and EDESC show competitive performance, with EDESC scoring slightly higher in $S_{eva}$ than IDEC. However, both are outperformed by TS-IDEC, which integrates convolutional feature extraction, dual-mode clustering, and composite evaluation into a unified pipeline. The superior DB score of TS-IDEC suggests that its clusters are not only compact but also well-separated, even in the presence of overlapping operational modes and nonstationary behavior. Moreover, TS-IDEC outperforms both IDEC and IDEC-conv1D, demonstrating that the proposed image-based transformation provides more effective feature representation for clustering time-series data.

Importantly, the optimal cluster number selected for TS-IDEC is four, as determined by maximizing $S_{eva}$, while other methods tend to prefer fewer clusters. This indicates that TS-IDEC is capable of capturing finer operational distinctions that may be missed by more rigid or assumption-heavy algorithms.

Overall, these findings validate the effectiveness and generalizability of TS-IDEC in real-world, label-free industrial settings. Its balanced performance across cohesion, separation, and compactness metrics makes it particularly suitable for applications requiring effective unsupervised learning, such as energy efficiency profiling and process optimization in manufacturing.

### 5.3. Cluster Explanation in Furnace Operations

After validating the performance of TS-IDEC through benchmark comparisons, this study analyzes the operational clusters discovered in the foundry melting dataset. These clustering outcomes reveal distinct temperature dynamics and energy-use profiles across different melting operations, offering insights into real-world industrial behavior.

Figure 6 presents the composite evaluation scores $S_{eva}$ achieved by TS-IDEC across different numbers of clusters. While the optimal result was observed at four clusters, local maxima at seven and twelve clusters suggest the presence of meaningful substructure in the data.

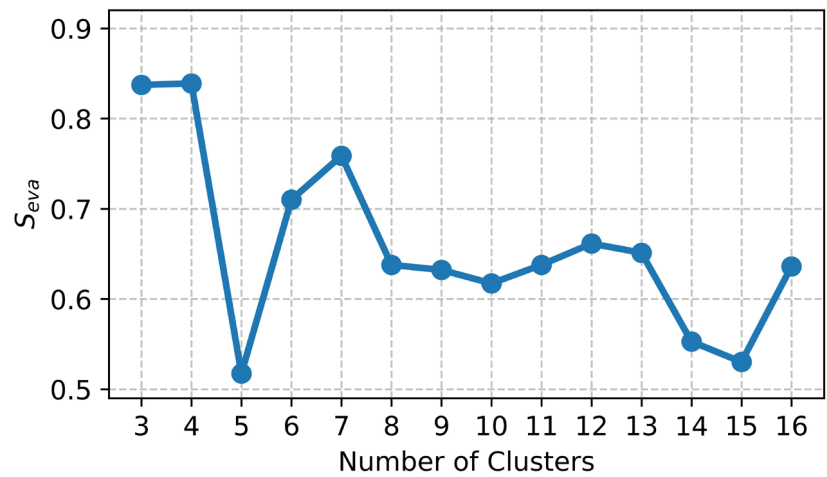

**Figure 6.** Normalized evaluation scores of different numbers of clusters for the TS-IDEC algorithm.

To further explore cluster explainability, this research examines three configurations, including four, seven, and twelve clusters, using both low-dimensional visualization and statistical summaries.

The t-SNE plots in Figure 7 show the latent representations of the time-series data projected into two dimensions for each cluster configuration. Each dot represents a single melting operation, and colors denote cluster membership.

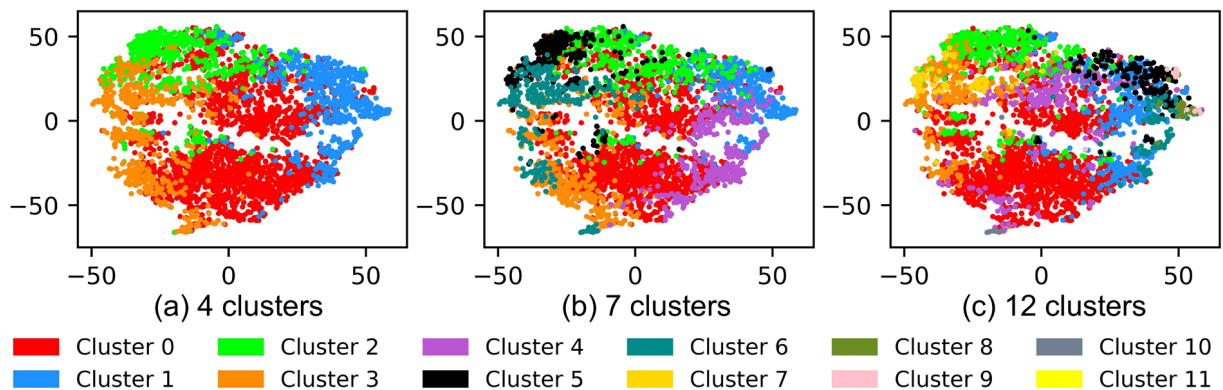

**Figure 7.** The t-SNE illustration of TS-IDEC clustering results in different numbers of clusters. Subfigures (**a–c**) depict the clustering outcomes for 4, 7 and 12 clusters, respectively. Each solid data point represents a time series projected into the t-SNE space, with colors indicating cluster assignments as defined in the legend.

The four-cluster solution (Figure 7a) shows relatively well-separated groups, but some overlap regions remain, indicating soft transitions or outliers. The seven-cluster solution (Figure 7b) provides finer granularity, with better intra-cluster compactness and inter-cluster separation. This configuration strikes a balance between model complexity and explainability. In contrast, the twelve-cluster result (Figure 7c)—obtained from a failed 12-cluster run where one cluster collapsed—shows excessive fragmentation and increased overlap, reducing explainability.

Table 5 provides a quantitative comparison of distinct clustering configurations for objective judgment. The dominant cluster ratio measures the share of samples in the largest cluster. Standard deviation reflects imbalance across cluster sizes. The number of small clusters indicates over-fragmentation, defined here as clusters containing less than 1% of the data.

**Table 5.** Comparison of distinct clustering solutions.

| k | Dominant Cluster Ratio (%) | Standard Deviation of Cluster Sizes | Number of Small Clusters (<1%) |
|---|---|---|---|
| 4 | 53.6 | 650.1 | 0 |
| 7 | 39.8 | 414.3 | 0 |
| 12 | 47.2 | 490.8 | 4 |

The results in Table 5 highlight clear trade-offs across the three candidate solutions. At k = 4, more than half of all samples (53.6%) fall into a single dominant cluster, with the highest standard deviation (650.1), indicating strong imbalance and loss of granularity. At the other extreme, k = 12 produces four very small clusters (<1% of data each), suggesting over-fragmentation and reduced stability. The intermediate solution, k = 7, achieves the lowest dominant cluster ratio (39.8%) and the lowest standard deviation (414.3), reflecting a more balanced distribution of samples across clusters without introducing spurious minor groups. These results support k = 7 as the most robust clustering solution, offering a compromise between compactness and granularity.

Figure 8 shows the average temperature profiles (cluster centers) for each configuration. These curves represent the dominant melting behavior of each cluster.

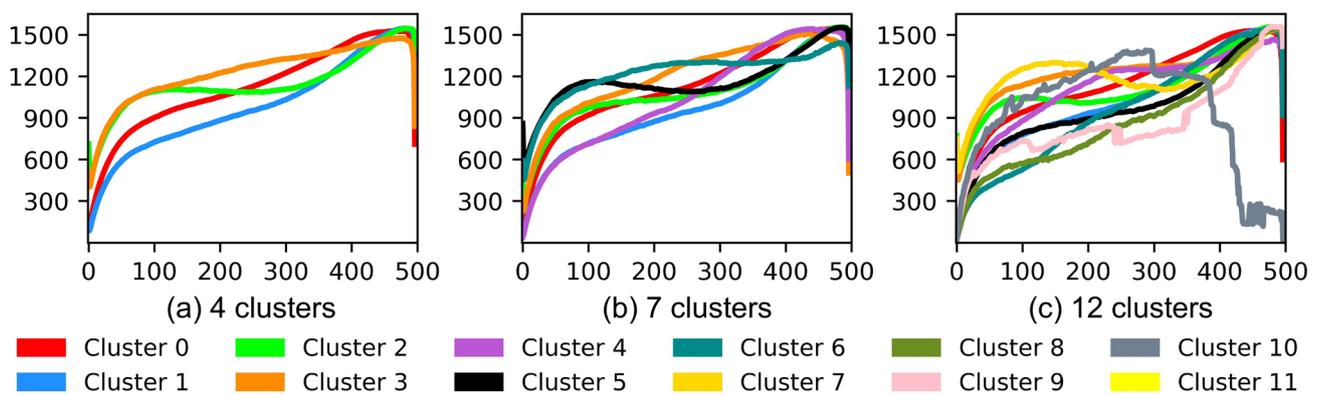

**Figure 8.** The cluster centers of TS-IDEC clustering with different numbers of clusters. Subfigures (**a**–**c**) depict the clustering outcomes for 4, 7 and 12 clusters, respectively. In each subfigure, the x-axis refers to the sequential data points, and the y-axis denotes furnace temperature (°C). Each curve represents a cluster center, with colors indicating cluster assignments as defined in the legend.

In all cases, the general structure of the melting operation is retained: a rapid heating phase followed by a slower convergence to a plateau. However, the clustering in Figure 8 reveals key variations in peak temperatures, ramp-up gradients, holding durations, and cooling profiles. These variations reflect underlying process differences such as material type, batch size, or operator behavior.

Table 6 summarizes key operational statistics for the seven-cluster configuration, which was selected for in-depth explainability based on the trade-off between granularity and clarity.

**Table 6.** Statistical Characteristics of Clusters (7-Cluster TS-IDEC Result).

| Attributes | Cluster 0 | Cluster 1 | Cluster 2 | Cluster 3 | Cluster 4 | Cluster 5 | Cluster 6 |
|---|---|---|---|---|---|---|---|
| Cardinality | 1562 | 488 | 466 | 440 | 353 | 313 | 305 |
| Average production time [s] | 4882.7 | 4340.5 | 4500.3 | 6043.2 | 4641.2 | 4963.7 | 7481.9 |
| Average weight [tonne] | 9.7 | 9.98 | 9.98 | 9.73 | 9.5 | 9.64 | 9.25 |
| Average electricity consumption [kWh] | 5068.4 | 5093.0 | 5260.1 | 5041.6 | 4853.6 | 5031.6 | 5054.5 |
| Average electricity consumption per unit [kWh/tonne] | 389.2 | 325.0 | 355.6 | 415.2 | 359.9 | 341.7 | 389.8 |

Cluster 0 is the most prevalent pattern, covering over one-third of the dataset. Cluster 6 is the least frequent but exhibits the longest average duration (7481.9 s), likely representing complex or delayed operations. Clusters 1 and 3 represent energy-use extremes: Cluster 1 shows the highest energy efficiency (325.0 kWh/tonne), while Cluster 3 consumes the most energy per unit (415.2 kWh/tonne), potentially indicating suboptimal or resource-intensive melting conditions.

These variations illustrate TS-IDEC's ability to uncover operationally significant regimes that can inform energy optimization, operator training, or process standardization. The separation of long-duration, inefficient batches from short, efficient ones could support targeted investigations or recommendations.

To further illustrate the temporal dynamics captured by the TS-IDEC model, detailed cluster-wise sequence plots are provided in Figures A1–A3 in the Appendix A. Each figure overlays all melting operation time series within a cluster configuration (4, 7, and 12 clusters), with the cluster centers highlighted in red. These visualizations reveal the degree of intra-cluster coherence and the diversity of temporal profiles across clusters. In particular, they confirm that TS-IDEC is capable of grouping sequences with similar thermal behaviors, while also maintaining separation between distinctly patterned operations. The increased fragmentation observed in the 12-cluster case (Figure A3) further supports the explainability argument made in favor of the 7-cluster configuration.

### 5.4. Robustness to Sliding Window Parameters

To evaluate the robustness of TS-IDEC with respect to its temporal transformation procedure, this paper conducted a sensitivity analysis on the sliding window parameters used to convert time-series sequences into grayscale matrices. Specifically, this paper varied both the window size and the stride length to observe their influence on clustering performance, as measured by the composite evaluation score $S_{eva}$.

Table 7 summarizes the six experimental configurations used in this analysis. Each configuration modifies either the window size or the stride, resulting in different matrix dimensions for input into the convolutional encoder.

**Table 7.** Sliding Window Parameter Settings.

| Config ID | Window Size | Stride | Resulting Matrix Shape |
|---|---|---|---|
| 1 (default) | 32 | 15 | 32 × 32 |
| 2 | 32 | 10 | 48 × 32 |
| 3 | 32 | 20 | 25 × 32 |
| 4 | 32 | 32 | 16 × 32 |
| 5 | 24 | 21 | 24 × 24 |
| 6 | 40 | 12 | 40 × 40 |

Configurations 1–4 vary the stride while keeping the window size fixed at 32, thus adjusting the degree of overlap between segments. Configurations 5 and 6 investigate the impact of varying the window size, which represents shorter or longer temporal patterns, respectively.

Figure 9 shows the $S_{eva}$ scores achieved by TS-IDEC under each configuration. The light blue bar indicates the default setting, while green and orange bars represent variations in stride and window size, respectively.

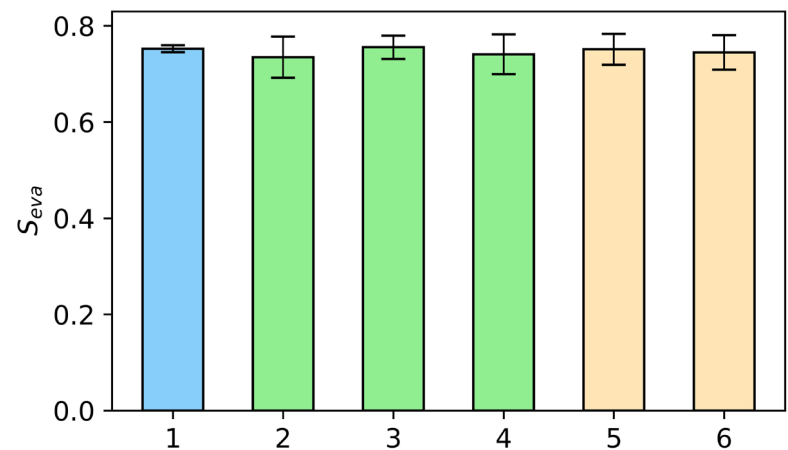

**Figure 9.** Evaluation scores of different experiments for sensitivity analysis. The x-axis corresponds to the experiment indices as presented in Table 6, while the y-axis indicates the evaluation score $S_{eva}$. The light blue bar represent the default configuration, the light green bars illustrate experiments analyzing the impact of stride, and the light orange bars depict experiments assessing the effect of varying the window size. The error bars represent the standard error of repetitions.

Across all six configurations, the model achieves consistently strong clustering performance, with only minor fluctuations in $S_{eva}$. The default setting (Config 1) yields slightly better average performance, but no statistically significant degradation is observed in any alternative configuration. In particular, the model remains stable even with minimal overlap (Config 4) or highly extended windows (Config 6), demonstrating robust generalization to variations in the temporal resolution of input data.

### 5.5. Design Justification Through Ablation Analysis

To validate the architectural and methodological choices in TS-IDEC, this paper conducted two ablation studies: (i) evaluating the contribution of the dual-mode clustering strategy, and (ii) assessing the impact of using the composite internal evaluation metric $S_{eva}$. These studies isolate the effects of key components and confirm their necessity for robust and high-quality clustering in complex industrial time-series data.

### 5.5.1. Impact of Dual-Mode Clustering

TS-IDEC combines soft clustering, based on Student's t-distribution, and hard clustering, based on k-means, to generate a final clustering outcome $C_{best}$. To assess the benefit of this hybrid strategy, this paper compared three configurations:

- Soft clustering only (TS-IDEC-soft);
- Hard clustering only (TS-IDEC-hard);
- Full TS-IDEC with selection between $C_1$ and $C_2$ Via the composite metric.

Figure 10 illustrates the normalized evaluation scores across different numbers of clusters for all three configurations, with subplots representing individual internal metrics and the final $S_{eva}$.

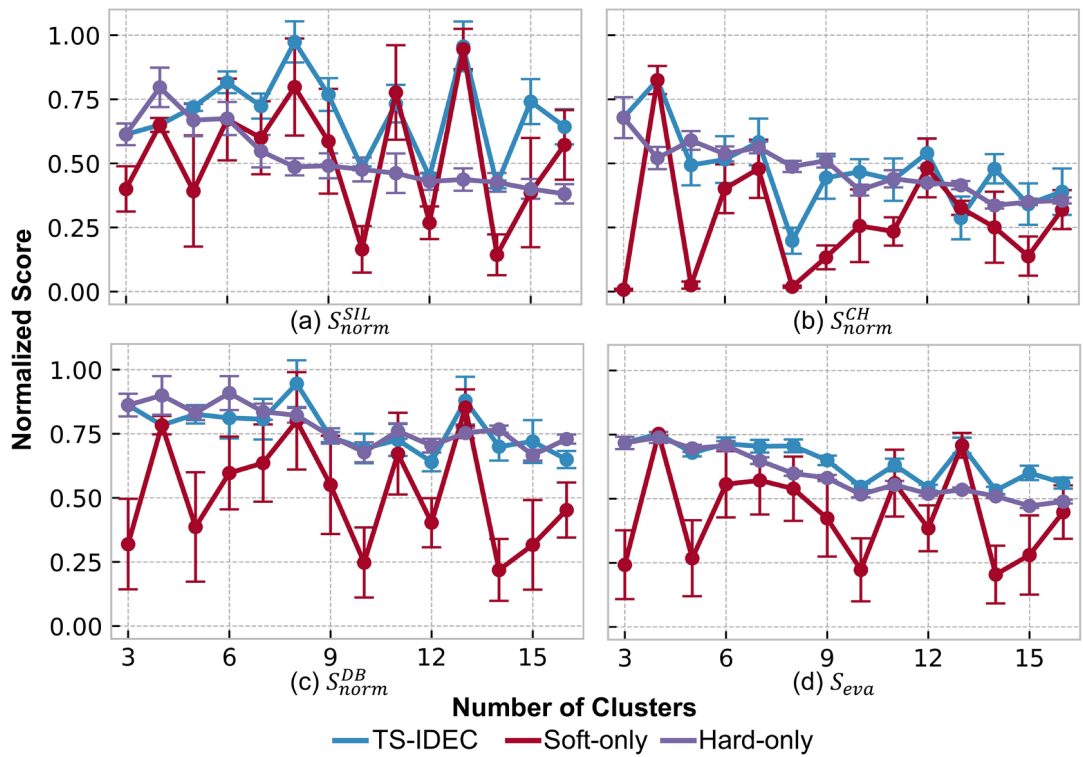

**Figure 10.** Evaluation scores of clustering ablation study. X-axes denote the number of clusters, and y-axes refer to the normalized evaluation scores. Sub-figures (**a–d**) represents the four evaluation scores $S_{norm}^{SIL}$, $S_{norm}^{CH}$, $S_{norm}^{DB}$ and $S_{eva}$, respectively. Three lines stand for the methods shown in the legend, respectively, and error bars are the standard deviation of repetitions.

The results indicate that hard clustering alone provides stable performance but struggles with soft transitions and noisy boundaries, which are common in real-world furnace operations. In contrast, soft clustering alone shows greater sensitivity and can capture nuanced behavior, but it suffers from convergence issues in some runs—sometimes collapsing into a single dominant cluster and significantly degrading performance.

### 5.5.2. Importance of Composite Evaluation Metric

To justify the use of the composite evaluation score $S_{eva}$ for model selection, this paper compared it against:

- Individual raw metrics: SIL, CH, and DB;
- Individual normalized metrics: $S_{norm}^{SIL}$, $S_{norm}^{CH}$, and $S_{norm}^{DB}$.

Table 8 summarizes the evaluation scores of all methods under these seven metrics, using the same optimal clustering result from each algorithm's best run.

**Table 8.** Comparison of clustering algorithms under different evaluation metrics.

| Methods | SIL | CH | DB | $S_{norm}^{SIL}$ | $S_{norm}^{CH}$ | $S_{norm}^{DB}$ | $S_{eva}$ |
|---|---|---|---|---|---|---|---|
| K-shape | **0.4640 ± 0.0** | 60.055 ± 0.0 | **1.5961 ± 0.0** | **0.9883 ± 0.0** | 0.0820 ± 0.0 | **0.9714 ± 0.0** | 0.6850 ± 0.0 |
| KM-DTW | 0.0438 ± 0.01080 | 111.38 ± 14.405 | 3.4444 ± 0.14373 | 0.7085 ± 0.03708 | 0.5026 ± 0.03122 | 0.4586 ± 0.02966 | 0.5566 ± 0.03052 |
| KM-softDTW | 0.1133 ± 0.01966 | 238.99 ± 22.794 | 2.9445 ± 0.24284 | 0.6477 ± 0.03654 | 0.4418 ± 0.02555 | 0.4727 ± 0.01966 | 0.5207 ± 0.00938 |
| IDEC | 0.1100 ± 0.01873 | 576.89 ± 54.273 | 2.1839 ± 0.11011 | 0.6209 ± 0.05454 | 0.9032 ± 0.05866 | 0.6882 ± 0.03852 | 0.7374 ± 0.04277 |
| IDEC-conv1d | 0.1056 ± 0.00447 | **584.11 ± 21.553** | 2.7173 ± 0.27621 | 0.6115 ± 0.01907 | **0.9166 ± 0.02443** | 0.5424 ± 0.05474 | 0.6902 ± 0.02199 |
| DTC | 0.0893 ± 0.00629 | 475.80 ± 36.108 | 2.8619 ± 0.12757 | 0.5320 ± 0.03071 | 0.7496 ± 0.05288 | 0.5081 ± 0.03712 | 0.5966 ± 0.03233 |
| EDESC | 0.1234 ± 0.00068 | 554.56 ± 3.677 | 2.1710 ± 0.01626 | 0.6922 ± 0.00395 | 0.8775 ± 0.00347 | 0.6815 ± 0.00397 | 0.7504 ± 0.00154 |
| TS-IDEC | 0.1141 ± 0.00678 | 501.46 ± 48.373 | 1.9183 ± 0.09290 | 0.6494 ± 0.02740 | 0.8246 ± 0.05510 | 0.7830 ± 0.03607 | **0.7523 ± 0.00724** |

The raw metrics provide inconsistent guidance. For example, k-Shape achieves the highest SIL and DB values but ranks among the lowest in CH, which favors models with better global separation. This inconsistency illustrates the metric conflict discussed in Section 3.5.

Normalized metrics improve comparability, but they remain incomplete indicators when used in isolation. The composite metric $S_{eva}$, by contrast, provides a more balanced and stable signal. TS-IDEC's leading score in $S_{eva}$, despite not ranking first in every individual metric, confirms its robust all-around performance. This result justifies the use of $S_{eva}$ as a selection criterion for final clustering output and demonstrates the advantage of multi-perspective internal validation in unsupervised industrial contexts.

## 6. Discussion

This section synthesizes the experimental findings with the broader literature on unsupervised time-series clustering, highlighting the practical relevance, theoretical contributions, and methodological advancements of the proposed TS-IDEC framework. The discussion is structured around two core aspects: (1) explainability and value of the discovered clusters in industrial settings, and (2) the framework's empirical performance and innovation relative to existing approaches.

### 6.1. Explanation of Discovered Operational Modes

The clusters generated by TS-IDEC capture distinct operational behaviors in industrial furnace melting. These clusters are not merely statistical groupings but align with real thermodynamic and operational characteristics, including variations in heating ramp rates, holding times, energy consumption, and production duration. Unlike threshold-based or supervised approaches, TS-IDEC uncovers latent process modes in an unsupervised manner, revealing structures that may not be explicitly encoded in domain knowledge.

For example, Cluster 1 represents the most energy-efficient mode, combining the shortest average production time with the lowest energy consumption per ton—an indicator of best-practice operations. By contrast, Cluster 3, although characterized by a smooth temperature profile, exhibits the highest energy use per unit, suggesting inefficient or manually adjusted processes. Cluster 5, marked by mid-process temperature dips, likely reflects operator interventions or atypical material compositions. Such cluster-specific insights extend prior work, such as the study in [8] that used structural entropy clustering for process monitoring, by offering a higher-resolution view of operational substructure.

Moreover, its ability to handle soft transitions, as demonstrated in the t-SNE projections and cluster center analyses (Figures 7 and 8), suggests that TS-IDEC is particularly well-suited for processes with high intra-mode variability and loosely defined operational boundaries. Full time-series overlays (Figures A1–A3 in Appendix A) further demonstrate consistency between individual sequences and their cluster centers, illustrating patterns such as dual-stage heating in Clusters 1 and 2 and irregular trajectories in Cluster 5. The strong alignment between data-driven clusters and physically plausible behaviors underscores TS-IDEC's value for unsupervised industrial diagnostics.

Beyond numerical clustering performance, TS-IDEC yields actionable insights for practice. The identified clusters provide operators with clear references for energy-optimal and inefficient behaviors, support targeted training, and inform engineers in refining control strategies or diagnosing anomalies. This grounding in domain relevance distinguishes TS-IDEC from abstract clustering frameworks. By extracting explainable clusters from noisy, unlabeled furnace data, TS-IDEC enables fine-grained understanding of melting patterns and operational variability. These capabilities position it as a scalable decision-support tool for energy-aware process monitoring and optimization in modern manufacturing environments.

### 6.2. Contributions Relative to Time-Series Clustering

TS-IDEC advances time-series clustering by integrating four innovations not previously combined in a single framework. Compared to foundational deep clustering models such as IDEC [10], DEC [9], and DTC [11], it introduces methods tailored to industrial, variable-length time-series data.

First, an overlapping sliding-window transformation converts univariate sequences into 2D matrix representations, enabling convolutional operations for spatial–temporal feature extraction. This preprocessing enhances robustness to local variations and allows convolutional learning (e.g., [43]), underutilized in prior unsupervised time-series clustering, to be effectively applied in industrial settings.

Second, a deep convolutional autoencoder (DCAE) extracts spatial–temporal features from these matrices, yielding better latent separation than fully connected or 1D convolutional models (e.g., IDEC-conv1D). This design leverages image-based representation [12] while extending them to unsupervised industrial contexts.

Third, TS-IDEC employs a dual-mode clustering strategy that integrates soft assignments (Student's t-distribution) with hard assignments (k-means). Ablation studies in Section 5.5.1 show this approach prevents cluster collapse, improves stability, and avoids the noise sensitivity of single-paradigm models, without increasing inference complexity.

Fourth, a composite evaluation score $S_{eva}$ combines normalized SIL, CH and DB indices through rank aggregation. This resolves inconsistencies among individual metrics [37,44], enabling balanced model selection that accounts for compactness, cohesion, and separation.

Collectively, these contributions establish TS-IDEC as a generalizable and scientifically rigorous framework for clustering industrial time series. The design addresses domain needs, such as explainability, robustness to variable operation lengths, and energy awareness, while filling methodological gaps in deep unsupervised learning. Its consistent top performance across all internal metrics confirms the value of the dual-mode clustering strategy, while sensitivity analyses in Figure 9 demonstrate robustness to window and stride parameters. These qualities make TS-IDEC reliable, explainable, and practical for deployment in large-scale, rapidly evolving manufacturing environments.

## 7. Conclusions

This paper presents TS-IDEC, a novel deep time-series clustering framework for uncovering hidden operational patterns in complex, unlabeled industrial datasets. By transforming univariate time series into grayscale matrices through overlapping sliding windows, the framework enables convolutional neural networks to extract spatial–temporal features. TS-IDEC integrates a deep convolutional autoencoder with a dual-mode clustering strategy, combining soft and hard assignments, and selects the final result Via a two-stage mechanism guided by a composite evaluation score.

To address limitations in existing unsupervised clustering evaluation methods, this paper proposes a data-driven composite metric, $S_{eva}$, which integrates normalized SIL, CH and DB indices. This metric reduces inconsistencies between individual scores and ensures stable model selection in the absence of ground truth. Comparative experiments demonstrated that TS-IDEC consistently outperforms classical methods (e.g., k-means-DTW, k-shape), deep clustering baselines (e.g., IDEC, DTC, EDESC), and convolutional variants (e.g., IDEC-conv1D).

Application to real-world furnace operations in a Nordic foundry confirmed the framework's ability to identify meaningful and explainable melting modes. Seven distinct clusters are discovered, revealing systematic differences in energy efficiency, process duration, and thermal dynamics. These insights enable benchmarking of operational performance, detection of inefficiencies, and formulation of data-driven strategies for process optimization, energy savings, and workforce training.

Scientifically, TS-IDEC advances unsupervised time-series clustering by addressing three major challenges: handling variable-length sequences, reducing instability in soft clustering, and providing a consistent evaluation criterion. The framework combines image-based representation learning, dual-mode clustering resilience, and reproducible evaluation, thereby improving rigor and reliability in unstructured industrial settings.

Practically, TS-IDEC generates domain-aligned, explainable insights without labeled data or manual feature engineering, supporting its deployment in smart manufacturing environments. It enables large-scale benchmarking across thousands of unlabeled cycles and highlights energy-optimal operating modes and intervention-heavy batches.

Despite its strengths, the approach remains sensitive to the selection of cluster numbers, underscoring the difficulty of modeling overlapping or ambiguous regimes. Future research could integrate recent deep learning–based time-series clustering approaches, including Transformer and attention-based architectures, both to establish a more comprehensive benchmark against state-of-the-art methods and to potentially advance clustering performance. Further automation of the windowing transformation, extension to multivariate and multimodal time series, and integration with self-supervised learning may enhance robustness and generalizability. Furthermore, future work should validate the proposed composite metric against datasets with ground truth to further establish its robustness and generalizability. Finally, coupling TS-IDEC with explainable AI methods could deepen interpretability by linking discovered clusters to underlying process parameters and enabling human-in-the-loop decision support.

**Author Contributions:** Conceptualization, Z.M., B.N.J. and Z.G.M.; methodology, Z.M., B.N.J. and Z.G.M.; software, Z.M.; validation, Z.M.; formal analysis, Z.M.; investigation, Z.M.; resources, B.N.J. and Z.G.M.; data curation, Z.M.; writing—original draft preparation, Z.M.; writing—review and editing, Z.M., B.N.J. and Z.G.M.; visualization, Z.M.; supervision, B.N.J. and Z.G.M.; project administration, B.N.J. and Z.G.M.; funding acquisition, Z.G.M. All authors have read and agreed to the published version of the manuscript.

**Funding:** This work was funded by the Energy Technology Development and Demonstration Programme (EUDP) in Denmark under the "Data-driven best-practice for energy-efficient operation of industrial processes—A system integration approach to reduce the CO2 emissions of industrial processes" funded by EUDP (Case no.64022-1051).

**Data Availability Statement:** The data that supports the findings of this study were obtained from an industrial partner and are subject to confidentiality agreements. As such, the original dataset cannot be shared publicly. Researchers interested in the methodology or derived results may contact the corresponding author for further discussion or guidance on reproducing the framework using their own data.

**Acknowledgments:** During the preparation of this work, the authors used *ChatGPT 4o* to improve the readability and language of the manuscript. After using this tool, the authors reviewed and edited the content as needed and take full responsibility for the content of the published article.

**Conflicts of interest:** The authors declare no conflicts of interest.

## Abbreviations

The following abbreviations are used in this manuscript:

| | |
|---|---|
| CAE | Convolutional Autoencoder |
| CH | Calinski–Harabasz Index |
| DB | Davies–Bouldin Index |
| DCAE | Deep Convolutional Autoencoder |
| DEC | Deep Embedded Clustering |
| DEETO | Deep Embedding with Topology Optimization |
| DTC | Deep Temporal Clustering |
| DTW | Dynamic Time Warping |
| EDESC | Efficient Deep Subspace Clustering |
| GAN | Generative Adversarial Network |
| IDEC | Improved Deep Embedded Clustering |
| IQR | Interquartile Range |
| KL | Kullback–Leibler |
| MSE | Mean Squared Error |
| SIL | Silhouette Score |
| TS-IDEC | Time-Series Image-based Deep Embedded Clustering |
| soft-DTW | Soft Dynamic Time Warping |

## Appendix A

Tables A1 and A2 present the parameter settings of the DCAE architecture in TS-IDEC and IDEC-conv1d, respectively.

**Table A1.** Parameter settings of the DCAE architecture used in TS-IDEC.

| Block Name | Layer Type | Kerner Size | Stride | Padding | Filters/Neurons | Activation |
|---|---|---|---|---|---|---|
| **Encoder** | | | | | | |
| conv1 | convolutional layer | 3 × 3 | 1 | 2 | 16 | ReLU |
| maxpool1 | maxpooling layer | 2 × 2 | 2 | 0 | N/A | N/A |
| conv2 | convolutional layer | 5 × 5 | 1 | 0 | 32 | ReLU |
| maxpool2 | maxpooling layer | 2 × 2 | 2 | 0 | N/A | N/A |
| conv3 | convolutional layer | 3 × 3 | 1 | 1 | 32 | ReLU |
| conv4 | convolutional layer | 3 × 3 | 1 | 1 | 64 | ReLU |
| fc1-en | fully connected layer | N/A | N/A | N/A | 2304 | ReLU |
| fc2-en | fully connected layer | N/A | N/A | N/A | 1028 | ReLU |
| fc3-en | fully connected layer | N/A | N/A | N/A | 512 | ReLU |
| **Latent space** | | | | | | |
| fc-latent | fully connected layer | N/A | N/A | N/A | 128 | ReLU |
| **Decoder** | | | | | | |
| fc3-de | fully connected layer | N/A | N/A | N/A | 512 | ReLU |
| fc2-de | fully connected layer | N/A | N/A | N/A | 1028 | ReLU |
| fc1-de | fully connected layer | N/A | N/A | N/A | 2304 | ReLU |
| deconv4 | deconvolutional layer | 3×3 | 1 | 1 | 32 | ReLU |
| deconv3 | deconvolutional layer | 3 × 3 | 1 | 1 | 32 | ReLU |
| upsample2 | upsampling layer | 2 × 2 | 2 | 0 | N/A | N/A |
| deconv2 | deconvolutional layer | 5 × 5 | 1 | 0 | 16 | ReLU |
| upsample1 | upsampling layer | 2 × 2 | 2 | 0 | N/A | N/A |
| deconv1 | deconvolutional layer | 3 × 3 | 1 | 2 | 1 | N/A |

**Table A2.** Parameter setting of the DCAE architecture of the IDEC-conv1d algorithm.

| Block Name | Layer Type | Kerner Size | Stride | Padding | Filters/Neurons | Activation |
|---|---|---|---|---|---|---|
| **Encoder** | | | | | | |
| conv1 | convolutional layer | 5 | 3 | 2 | 8 | ReLU |
| avgpool1 | average pooling layer | 2 | 2 | 0 | N/A | N/A |
| conv2 | convolutional layer | 3 | 1 | 1 | 16 | ReLU |
| avgpool2 | average pooling layer | 2 | 2 | 0 | N/A | N/A |
| conv3 | convolutional layer | 3 | 1 | 1 | 16 | ReLU |
| fc1-en | fully connected layer | N/A | N/A | N/A | 672 | ReLU |
| fc2-en | fully connected layer | N/A | N/A | N/A | 512 | ReLU |
| fc3-en | fully connected layer | N/A | N/A | N/A | 256 | ReLU |
| **Latent space** | | | | | | |
| fc-latent | fully connected layer | N/A | N/A | N/A | 64 | ReLU |
| **Decoder** | | | | | | |
| fc3-de | fully connected layer | N/A | N/A | N/A | 256 | ReLU |
| fc2-de | fully connected layer | N/A | N/A | N/A | 512 | ReLU |
| fc1-de | fully connected layer | N/A | N/A | N/A | 672 | ReLU |
| deconv3 | deconvolutional layer | 3 | 1 | 1 | 16 | ReLU |
| upsample2 | upsampling layer | 2 | 2 | 0 | N/A | N/A |
| deconv2 | deconvolutional layer | 3 | 1 | 1 | 8 | ReLU |
| upsample1 | upsampling layer | 2 | 2 | 0 | N/A | N/A |
| deconv1 | deconvolutional layer | 5 | 3 | 2 | 1 | N/A |

Figures A1–A3 demonstrate the results of TS-IDEC clustering with 4, 7, 10 clusters, respectively.

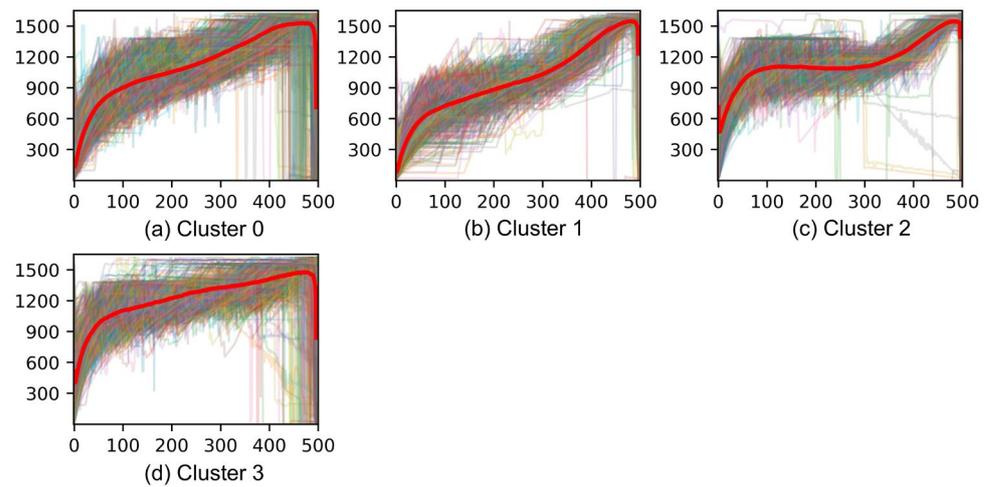

**Figure A1.** Illustration of TS-IDEC clustering results with 4 clusters. Subfigures (**a**–**d**) show the time-series data for clusters 0–3, respectively. In each subfigure, the bold red curve represents the cluster center, while the remaining curves depict individual time series within the cluster. The x-axis refers to the sequential data points, and the y-axis denotes furnace temperature (°C).

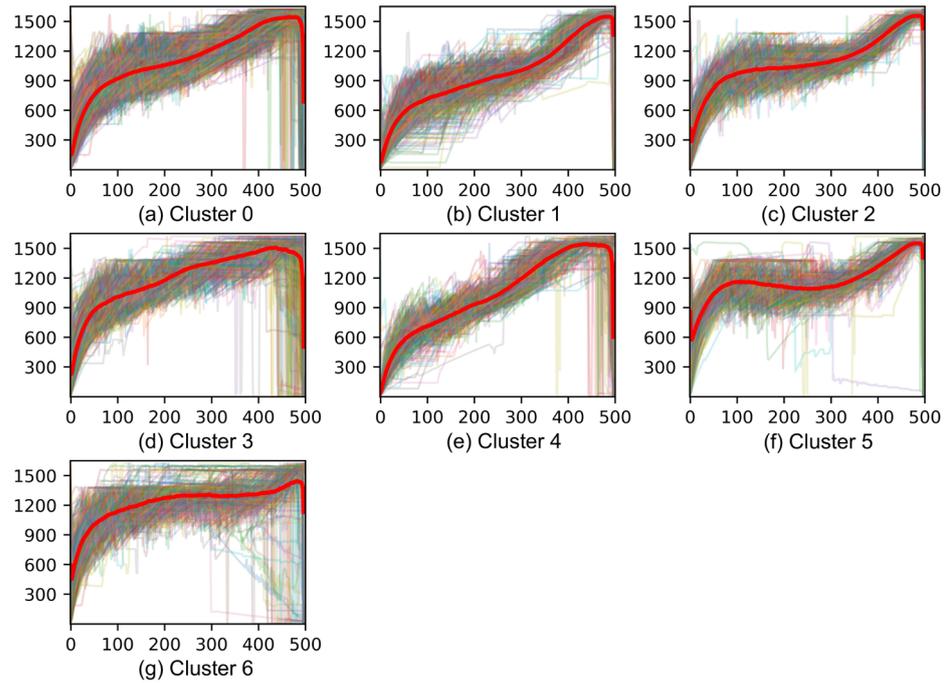

**Figure A2.** Illustration of TS-IDEC clustering results with 7 clusters. Subfigures (**a–g**) show the time-series data for clusters 0–6, respectively. In each subfigure, the bold red curve represents the cluster center, while the remaining curves depict individual time series within the cluster. The x-axis refers to the sequential data points, and the y-axis denotes furnace temperature (°C).

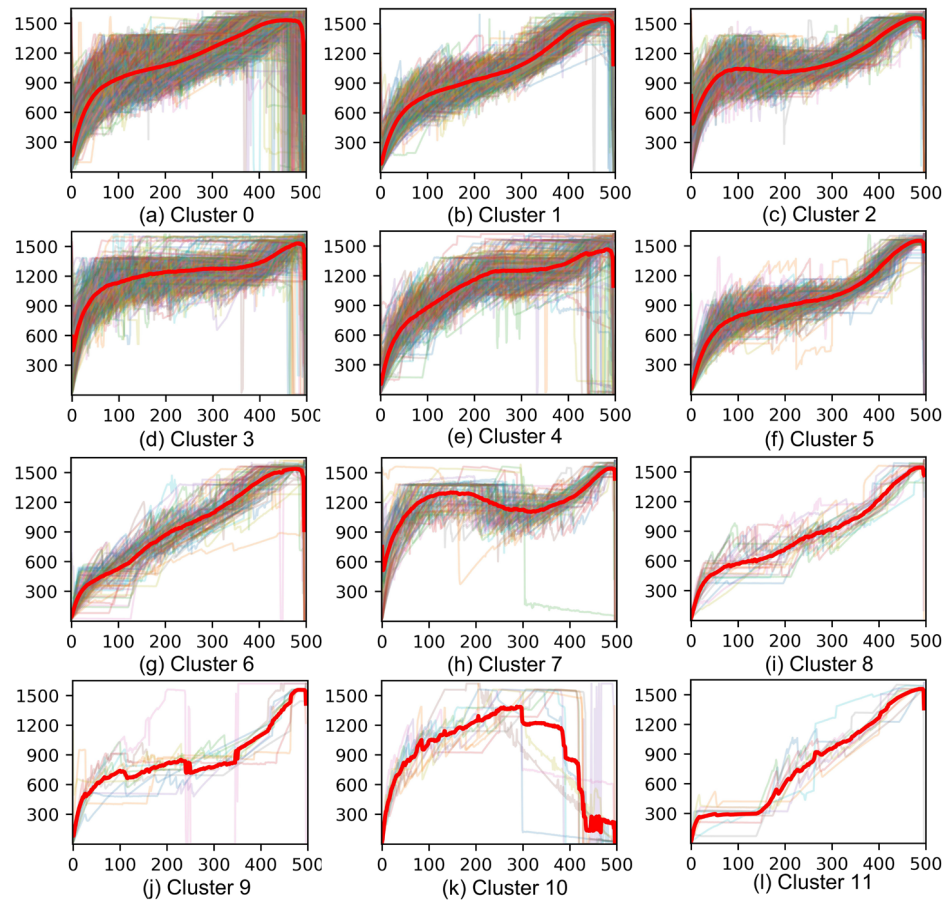

**Figure A3.** Illustration of TS-IDEC clustering results with 12 clusters. Subfigures (**a–l**) show the time-series data for clusters 0–11, respectively. In each subfigure, the bold red curve represents the cluster

center, while the remaining curves depict individual time series within the cluster. The x-axis refers to the sequential data points, and the y-axis denotes furnace temperature (°C).